\documentclass[a4paper, 10 pt, conference]{ieeeconf}  

\IEEEoverridecommandlockouts                              

\usepackage{graphics} 
\usepackage{epsfig} 

\title{\LARGE \bf
Two-Stream Networks for Lane-Change Prediction of Surrounding Vehicles}

\author{David Fern\'{a}ndez-Llorca$^{1}$, Mahdi Biparva$^{2}$, Rub\'{e}n Izquierdo-Gonzalo$^{1}$ and John K. Tsotsos$^{2}$
\thanks{$^{1}$D. F. llorca and R. Izquierdo are with the Computer Engineering Department, University of Alcalá, Alcalá de Henares, Madrid, Spain
        {\tt\small david.fernandezl@uah.es}}%
\thanks{$^{2}$M. Biparva and J. K. Tsotsos are with the Department of Electrical Engineering and Computer Science, York University, Toronto, ON M3J 1P3, Canada
        {\tt\small tsotsos@eecs.yorku.ca}}%
}

\begin{document}

\maketitle
\thispagestyle{empty}
\pagestyle{empty}

\begin{abstract}
In highway scenarios, an alert human driver will typically anticipate early cut-in and cut-out maneuvers of surrounding vehicles using only visual cues. An automated system must anticipate these situations at an early stage too, to increase the safety and the efficiency of its performance. To deal with lane-change recognition and prediction of surrounding vehicles, we pose the problem as an action recognition/prediction problem by stacking visual cues from video cameras. Two video action recognition approaches are analyzed: two-stream convolutional networks and spatiotemporal multiplier networks. Different sizes of the regions around the vehicles are analyzed, evaluating the importance of the interaction between vehicles and the context information in the performance. In addition, different prediction horizons are evaluated. The obtained results demonstrate the potential of these methodologies to serve as robust predictors of future lane-changes of surrounding vehicles in time horizons between 1 and 2 seconds.

\end{abstract}

\section{INTRODUCTION}
One of the closest and most plausible scenarios in the adoption of the autonomous vehicles is autonomous navigation at SAE L3 (chauffeur) or L4 (autopilot) on highways, both for passenger and freight transport. The most advanced automation systems to date are the Highway Chauffeur (HC) and the Highway Autopilot (HA), which includes the management of complex maneuvers such as deciding to change lanes to overtake, enter a slower lane or even exit the highway. HC is mostly considered as L3 and HA as L4\cite{ERTRAC2019}. In these systems, the most critical, and challenging, highway scenarios are the cut-in and cut-out ones, specially for high speeds. In the cut-in scenario, a car from one of the adjacent lanes merges into the lane just in front of the ego car. In the cut-out scenario, a car in front leaves the lane abruptly to avoid a slower vehicle, or even stopped, ahead. Since 2018, the performance of these assistance or chauffeur commercial systems operating under these two critical traffic scenarios is being tested by Euro NCAP \cite{EURONCAP2018}.

An alert driver will typically anticipate early cut-in and cut-out maneuvers using only visual cues, reduce speed accordingly or even change lanes through the use of the steering wheel. An automated system must also be able to anticipate these situations at an early stage. To do so, it is necessary to endow new automated systems with the ability of predicting the motions of surrounding vehicles, such as lane-keeping and lane-change.  

To deal with lane-change prediction of surrounding vehicles, in this paper we pose the problem as an action recognition problem using visual information from cameras. The idea behind our proposal is to use the same source of information (visual cues) and the same type of approach (action recognition) that drivers use to anticipate these maneuvers. 

\begin{figure}[!t]
	\centering	
	\includegraphics[width=0.48\textwidth]{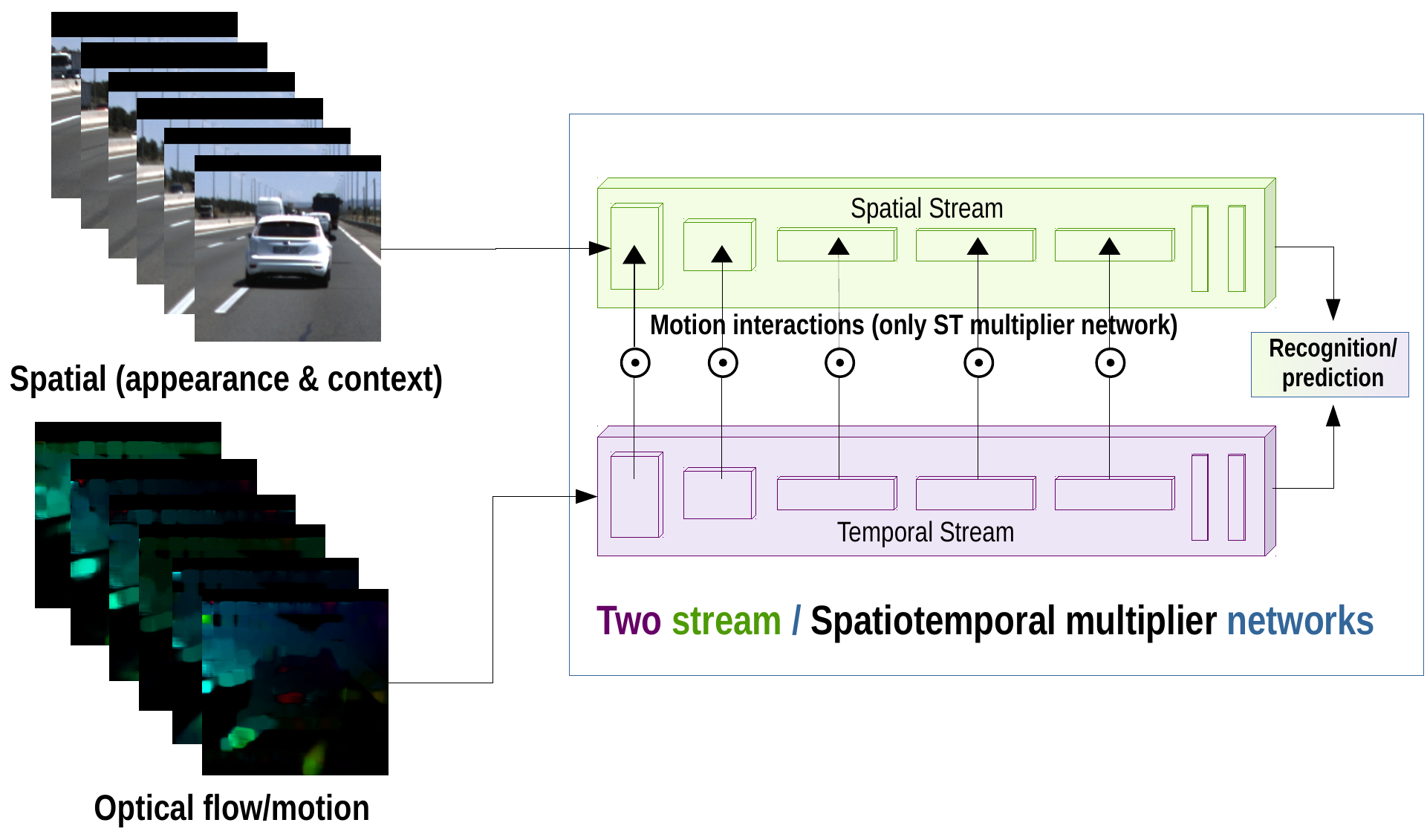}
	\caption{Overview of the proposed video action recognition approaches for lane change recognition and prediction of surrounding vehicles, including Two-Stream Network and Spatiotemporal Multiplier Network.}
	\label{fig:overview}
\end{figure}

Significant progress has been made in video-based human action recognition and prediction during the last years \cite{Kong2018}. Action recognition and prediction involves managing spatial and temporal information (sequence of images). Among the different methodologies, in this paper, we study \textbf{Two-Stream Convolutional Networks} \cite{Simonyan2014} and \textbf{Spatiotemporal Multiplier Networks} \cite{Feichtenhofer2017} approaches (see Figure \ref{fig:overview}) using The PREVENTION dataset \cite{prevention_dataset} to train and validate them. 

\section{Related Work}
\label{sec:soa}
Most of the available work on lane-change recognition and prediction focuses on lane-changes of the ego-vehicle. However, the nature of the problem is considerably different when we focus on lane-changes of surrounding vehicles, so we limit our analysis of lane-change detection of other vehicles, within the context of the highway scenario. 

\subsection{Input variables}
Most of the previous works are based on the use of physical variables that define the relative dynamics of the vehicle with other vehicles and with its environment \cite{Kasper2012}, \cite{Graf2013}, \cite{Schlechtriemen2014}, \cite{Liu2014}, \cite{Schlechtriemen2015}, \cite{Yoon2016}, \cite{Bahram2016},\cite{Izquierdo2017}, \cite{Yao2017}, \cite{Lee2017}, \cite{Deo2018}, \cite{Deo2018b}, \cite{patel2018predicting}, \cite{Li2019e}, \cite{Kruger2019}, including lateral and longitudinal distances, velocity, acceleration, timegap, heading angle and yaw rate. 

Context cues are also introduced, including road-level features such as the curvature and speedlimit \cite{Schlechtriemen2014}, \cite{Schlechtriemen2015}, \cite{Li2019e}, distance to the next highway junction \cite{Bahram2016}, number of lanes \cite{patel2018predicting}, etc., as well as lane-level features such as type of lane marking or the distance to lane end \cite{Bahram2016}. 

The number of proposals making use of appearance features is surprisingly low, especially considering that human drivers do not use the physical variables mentioned above to anticipate lane changes from other vehicles but visual cues. In \cite{Li2019e}, two variables manually selected from the appearance, i.e., state of turn indicators and state of brake indicators, are used. In our previous work \cite{Izquierdo2019} regions of interest (ROIs) are generated for each vehicle detection, including local information around the vehicle, and appearance features are extracted using a GoogLeNet pre-trained on ImageNet.  

\subsection{Methodologies}
As suggested by \cite{Lefevre2014} vehicle motion modeling and prediction approaches can be classified into three different levels: physical-based, where predictions only depend on the laws of physics, maneuver-based, where the future motion of a vehicle depends on the driver maneuver, and  intention-aware, where predictions take into consideration inter-dependencies between vehicles. 
 
Some proposals are intention-aware in their nature. For example, by using graphical models such as Bayesian Networks \cite{Kasper2012}, \cite{Bahram2016}, \cite{Li2019e} or Structural Recurrent Neural Networks \cite{patel2018predicting}, or by using convolutional social pooling in an LSTM encoder-decoder architecture \cite{Deo2018b}. However, in most cases, inter-dependencies between vehicles are modeled by extracting relative physical features \cite{Schlechtriemen2014}, \cite{Schlechtriemen2015}, \cite{Yao2017}, \cite{Deo2018} or by generating compact representations that encode the relative positions of all vehicles on the scene \cite{Lee2017}, \cite{Izquierdo2019}. Some works do not take into consideration the interaction between vehicles \cite{Graf2013}, \cite{Liu2014}, \cite{Yoon2016}, \cite{Li2019itsc}, \cite{Kruger2019}. 

Many approaches to lane-change recognition and prediction address the problem using generative-based solutions, including Na\"{i}ve Bayes Classifiers \cite{Schlechtriemen2014}, Bayesian Networks \cite{Kasper2012}, \cite{Bahram2016}, \cite{Li2019e}, and Hidden Markov Models \cite{Liu2014}. Others make use of discriminative solutions such as case-based reasoning \cite{Graf2013}, Random Decision Forest \cite{Schlechtriemen2015}, traditional Neural Networks \cite{Yoon2016}, \cite{Izquierdo2017}, Support Vector Machines \cite{Izquierdo2017}, \cite{Yao2017}, \cite{Li2019itsc}, Gaussian Process Neural Networks \cite{Kruger2019}, and feedforward Convolutional Neural Networks \cite{Lee2017}, \cite{Izquierdo2019}. Finally, some other approaches are based on the use of Recurrent Networks including vanilla LSTM \cite{Izquierdo2019} and LSTM encoder-decoder \cite{Deo2018} and multi-modal \cite{Deo2018b} architectures. 

\subsection{Datasets}

Two type of recording setups are usually proposed depending on the location of the sensors. First, we have datasets captured from the infrastructure using cameras installed on buildings, such as NGSIM HW101 \cite{Ngsim101dataset} or NGSIM I-80 \cite{NgsimI80dataset} datasets, or cameras on-board drones, such as HighD \cite{highddataset}, inD \cite{inddataset} or INTERACTION \cite{interactiondataset} datasets. Although these datasets are very valuable for understanding and assessing the motion and behavior of vehicles and drivers under different traffic scenarios, they are not fully applicable for on-board detection applications. 

Second, other datasets provide road data with sensors on-board vehicles. In this line, the PKU dataset \cite{pkudataset} contains 170 minutes of data gathered using a vehicle equipped with 4 2D-LiDARs covering a region of 40 meters around the vehicle (road lane markings, number of road lanes, or the relative positioning of the ego-vehicle are not provided). The ApolloScape dataset \cite{apolloscapedataset} provides data obtained in urban environments from 4 cameras and 2 Laser scanners using a vehicle driving at 30 km/h. It does not contain radar data, making detections more sensitive to failure in adverse weather conditions, and it does not provide labeled tracking information (IDs and tracklets) for all detected objects. Recently, in 2019, the PREVENTION dataset \cite{prevention_dataset} was released containing data from 3 radars, 2 cameras and 1 LiDAR, covering a range of up to 80 meters around the ego-vehicle. Road lane markings are included and the final position of the vehicles is provided by fusing data from the three type of sensors. 

\begin{figure}[!t]
	\centering	
	\includegraphics[width=0.45\textwidth]{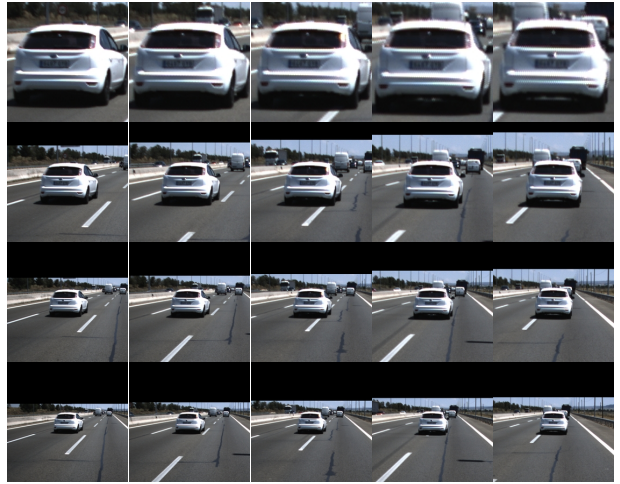}
	\caption{ROI sizes. From upper row to lower row: x1, x2, x3 and x4. The vehicle is always centered. Zero-padding is applied when needed.}
	\label{fig:rois}
\end{figure}
\begin{figure}[!ht]
	\centering	
	\includegraphics[width=0.47\textwidth]{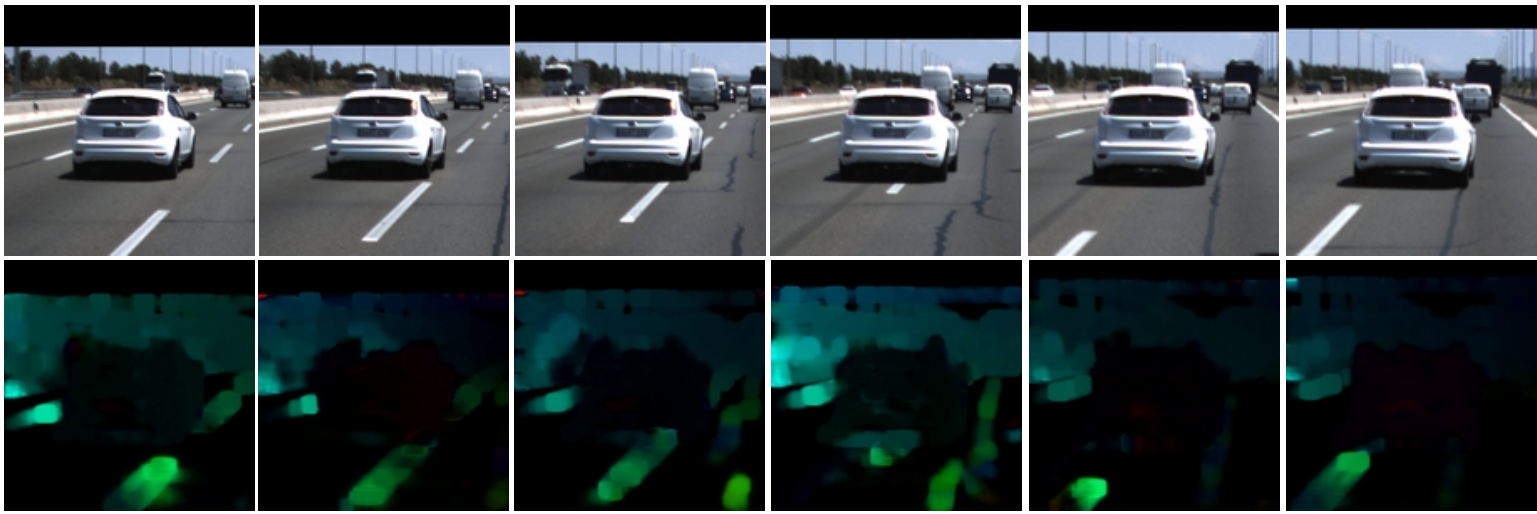}
	\caption{Example of dense optical flow computation.}
	\label{fig:of}
\end{figure}

\section{Problem Formulation}
\label{sec:problem}

\begin{figure*}[!ht]
	\centering	
	\includegraphics[width=0.97\textwidth]{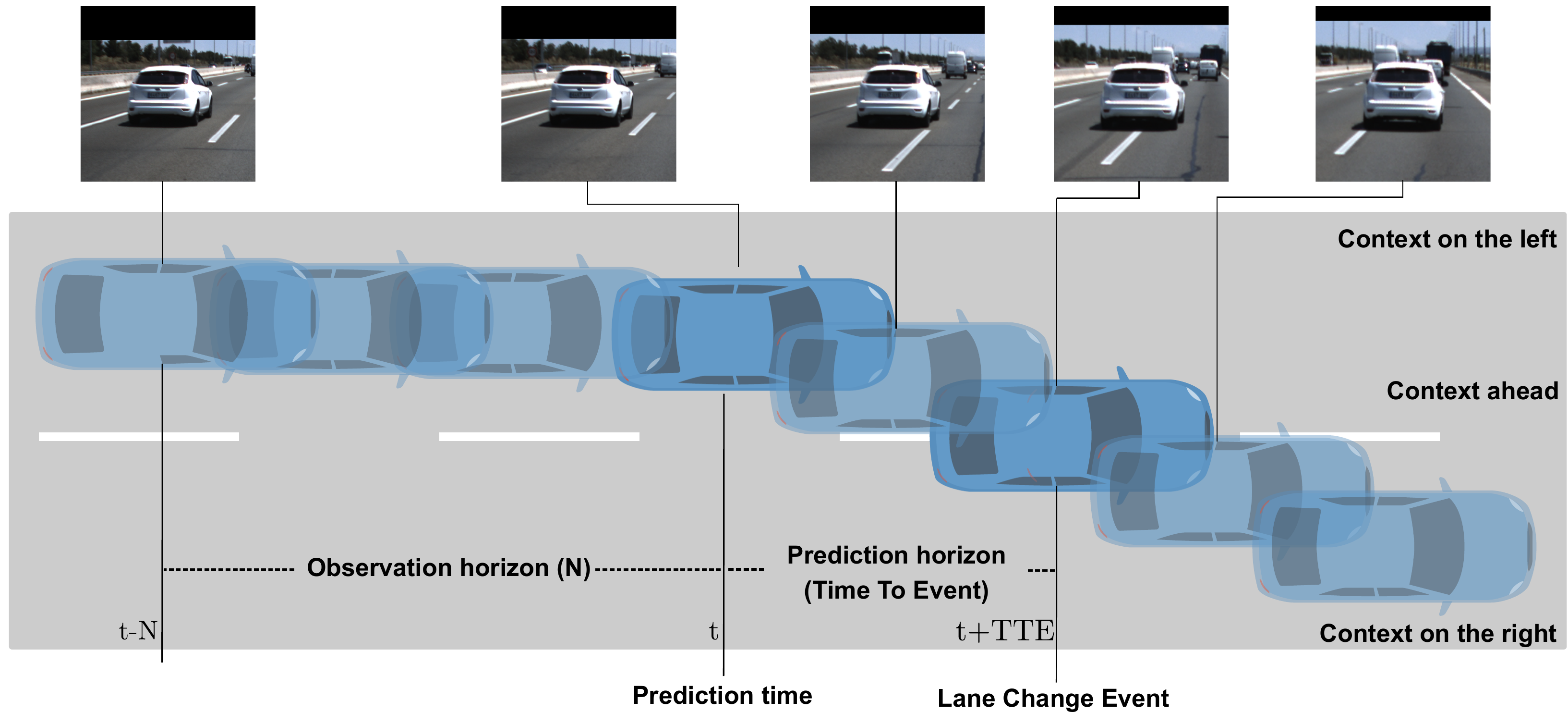}
	\caption{Problem formulation: observation horizon (N), and time to event (TTE). The lane change event is labeled as the frame where the  middle of the rear bumper is located just over the lane markings. This is the criterion established in PREVENTION dataset \cite{prevention_dataset}. }
	\label{fig:problem}
\end{figure*}

We define lane change prediction as a multi-classification problem in which the goal is to recognize  whether a vehicle $i$ will make a left or right lane-change or remain in its lane given the observed context up to some time $N$. The prediction relies on visual cues that are computed from regions of interest (ROIs) extracted from the contour labels provided in the PREVENTION dataset. Four different ROI sizes are considered: $\times 1$, $\times 2$, $\times 3$ and $\times 4$ the size of the square bounding box around the vehicle contour (see Figure \ref{fig:rois}). Zero-padding is used when the ROI exceeds the limits of the image. The size of the ROI modulates the amount of context information being considered in the input data stream. Thus, $\times 1$  mostly contains information related with the vehicle appearance, while $\times 4$ incorporates a large amount of front and side context information. For ROI sizes greater than $\times 2$, the approach can be considered as interaction-aware.

Since the vehicle is always centered in the ROI, dense optical flow (from the motion stream) should be interpreted as a way of measuring the movement of the context (infrastructure and other vehicles) around the detected vehicle. As shown in Figure \ref{fig:of}, the optical flow is low in the region where the vehicle is, while it is more predominant around it.

As can be seen in Figure \ref{fig:problem}, the lane-change event is defined as the time when the center of the rear bumper is just above the lane markings. The observation horizon or time window will contain a set of $N$ images that will be stacked according to the activity recognition method used. We will examine the effects of time to event (prediction horizon) and observation duration ($N$) on the accuracy of lane-change classification (when $TTE = 0$) and prediction (when $TTE>0$).


\section{Video activity recognition and prediction}
\label{sec:activity}
The sequence of stacked images or regions of interests, can naturally be decomposed into spatial and temporal components. The spatial part, in the form of individual region appearance, carries information about the vehicle itself (e.g., light indicators or brake lights) and the context around it (road, lane markings and surrounding vehicles). The temporal part, in the form of motion across frames, conveys the movement of the observer (onboard camera) w.r.t. to the road, and the surrounding vehicles. In order to handle a canonical view for the motion stream, all the regions are generated around the contour of the vehicle so the vehicle is always centered in the region of interest (the size will vary depending on the relative distance w.r.t. the ego vehicle). We consider two video activity recognition approaches: Disjoint Two-Stream Convolutional Networks \cite{Simonyan2014} and Spatiotemporal Multiplier Networks \cite{Feichtenhofer2017}.

\begin{figure*}[!ht]
	\centering	
	\includegraphics[width=0.99\textwidth]{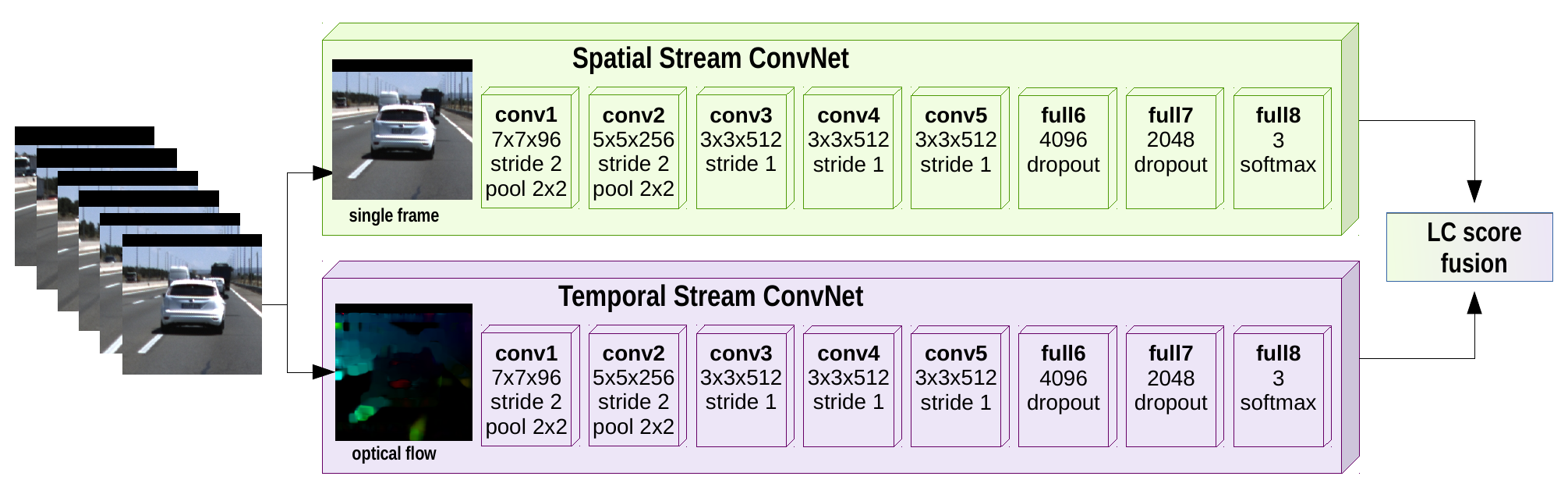}
	\caption{Disjoint two-stream architecture for lane change classification and prediction.}
	\label{fig:disjoinConvNet}
\end{figure*}

\subsection{Disjoint Two-Stream Convolutional Networks}
A two-stream ConvNet architecture which incorporates and fuses spatial and temporal information is defined. The structure of the ConvNets for both streams is the same, including 5 convolutional layers and 3 fully connected layers, with the parameters depicted in Figure \ref{fig:disjoinConvNet}. The last fully connected layer is defined with 3 outputs regarding the three classes defined: left lane change (LLC), right lane change (RLC), and no lane change (NLC). 

The dense optical flow is computed using polynomial expansion \cite{Farneback2003}. The spatial stream ConvNet is pre-trained using ImageNet and the temporal ConvNet using multi-task learning using UCF-101 and HMDB-51. All hidden layers use the rectification (ReLU) activation function. Max-pooling is performed over $3\times 3$ spatial windows with stride 2. 

\subsection{Spatiotemporal Multiplier Networks}

The original two-stream architecture only allows the two processing streams (spatial and motion) to interact via late fusion of their respective softmax predictions. This way, the architecture does not support the learning of truly spatiotemporal features, since the loss of both streams is backpropagated independently without any type of interaction. Learning spatiotemporal features requires the appearance and motion paths to interact earlier on during the forward pass. This interaction can be relevant for the classification and prediction of lane change maneuvers that have similar appearance or motion patterns and can only be inferred by the combination of two (e.g., vehicles that do not change lanes but have their turn indicators on). To address this limitation, it is possible to inject cross-stream residual connections using Residual Networks (ResNets) \cite{He2016} as the general architecture for the spatial and the temporal streams.

In \cite{Feichtenhofer2017}, different cross-stream connections were studied, including two types of connections (direct or into residual units), two fusion functions (additive or multiplicative), and different streams directions (unidirectional from the motion into the appearance, conversely and bidirectional), being the multiplicative residual connection from the motion path into the appearance stream the one providing the superior performance. 

\begin{figure}[!ht]
	\centering	
	\includegraphics[width=0.35\textwidth]{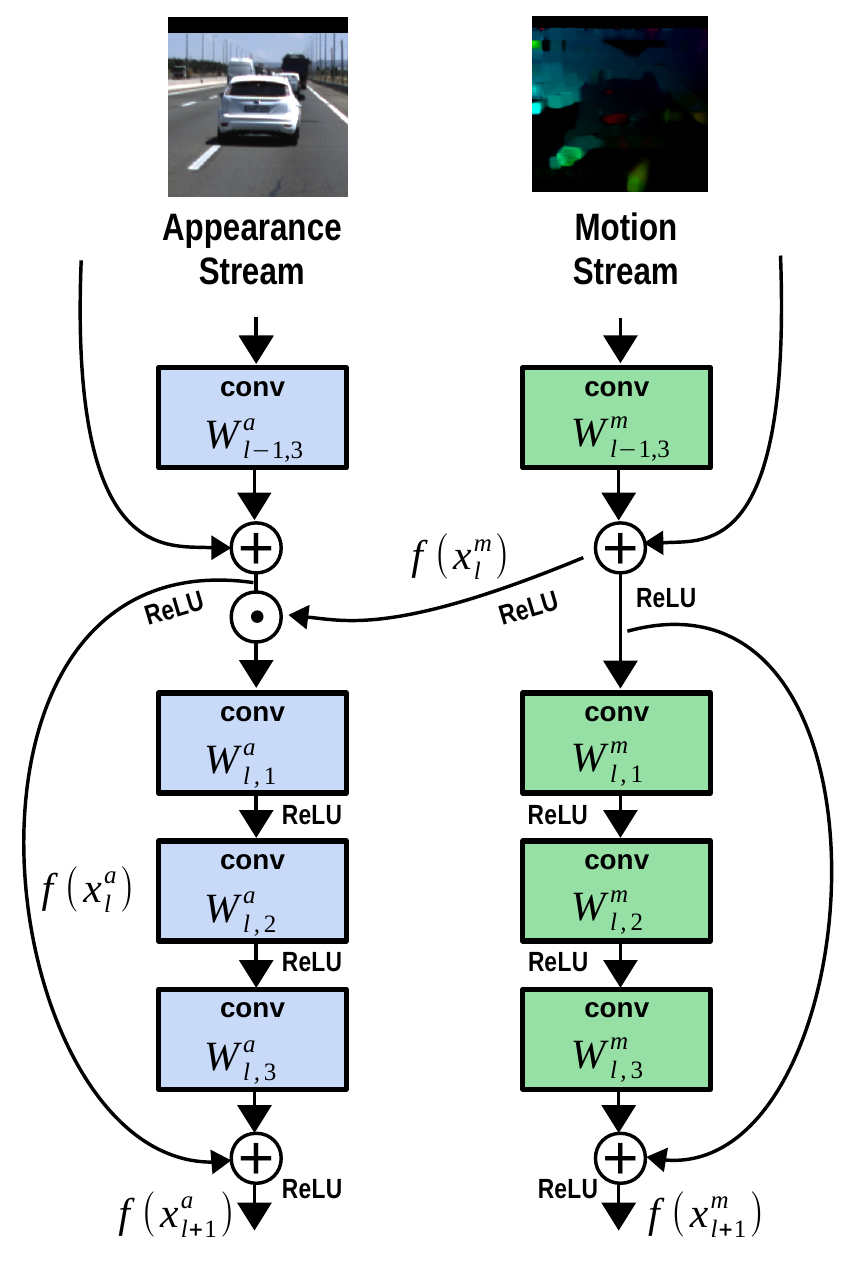}
	\caption{Multiplicative residual gating from the motion stream to the appearance stream.}
	\label{fig:stx}
\end{figure}

As can be observed in Figure \ref{fig:stx}, the multiplicative interaction can be formulated as:

\begin{equation}
\hat{x}^a_{l+1}=f(x_l^a)+\mathcal{F}\Big(x_l^a\odot f(x_l^m),W_l^a \Big)
\end{equation} 
where $x_l^a$ and $x_l^m$ are the inputs of the $l$-th layers of the appearance and motion paths respectively, while $W_l^a$ represents the weights of the $l$-th layer residual unit in the appearance stream and $\odot$ corresponds to elementwise multiplication. 
 
Better temporal support is also provided by injecting 1D temporal convolutions layers into the network \cite{Feichtenhofer2017}. ResNet50 model is used for both streams, including batch normalization and ReLU activation function after each convolutional block. 

\subsection{Recognition \& Prediction}
The proposed two-stream architectures have been historically applied to perform activity recognition from video sequences (e.g., human activity recognition), i.e., using a sequence of images (from $t-N$ frames up to $t$) to perform the recognition of the activity taking place in the video at time $t$. In order to perform prediction (at $t+TTE$), we define the target class (none, left or right) that will take place in a future time horizon given by $TTE$ as the desired class at $t$. In other words, as can be observed in Figure \ref{fig:problem}, we can consider that the system is performing lane change recognition or classification at time $t$ when $TTE=0$. In the same way, the system will be predicting the lane changes when $TTE>0$ (in frames). Taking into account that the sampling frequency of PREVENTION dataset is $10$Hz, we define the following scenarios:

\begin{itemize}
\item $TTE=0$ frames: lane-change classification at time $t$.
\item $TTE=10$ frames: lane-change prediction 1 second ahead ($t+10$). 
\item $TTE=20$ frames: lane-change prediction 2 seconds ahead ($t+20$). 
\end{itemize}


\section{Experimental Results}
\label{sec:results}

\subsection{Dataset description}
Table \ref{table:dataset} summarizes the details of the dataset. The input size for both streams is $112\times112$. The $85\%$ of the samples are used for training and the remaining $15\%$ for validation. 

\begin{table}[h!]
\centering
\begin{tabular}{|c|c|c|c|} 
  \hline
 & NLC & LLC & RLC \\
 \hline
 \# of sequences & 3110 & 342 & 438 \\ 
 avg. \# of frames & 50.9 & 96.8 & 80.1 \\ 
 \hline
\end{tabular}
\caption{Main stats of the dataset. NLC/LLC/RLC: no/left/right lane-change.}
\label{table:dataset}
\end{table}

\subsection{Evaluation parameters}
The following parameters have been evaluated during the experiments:

\begin{itemize}
\item ROI sizes: x1, x2, x3 and x4.
\item Observation horizon: 20 frames (2 seconds), 30 frames (3 seconds) and 40 frames (4 seconds).
\item Time-to-event (prediction horizon): 0 (no prediction), 10 (1 second) and 20 (2 seconds).
\end{itemize}

\subsection{Metrics}
As a multi-class problem (with 3 classes), we have considered the accuracy as the main variable to assess the performance of the two evaluated methods and the corresponding parameters, i.e., the number of true positives for the three classes divided by the total number of samples.

\subsection{Lane change classification results}
In Table \ref{table:1} we depict the accuracy of the two action recognition approaches over the validation set, i.e., with $TTE=0$. Regarding the ROI sizes we can state the following conlusions. By using just the ROI fitted to the bounding box, the results are surprisingly reasonable, considering that almost no context and interaction are available. In general, the higher the ROI size, the better the accuracy. However for observation horizon of 40 frames, adding more context from x3 to x4 decreases the performance. This can be explained by the fact that a larger observation horizon already incorporates more context into the spatial and motion streams.

For observation horizons of 20 and 30 frames, the simpler disjoint two-stream network offers better results than the spatiotemporal multiplier network. However, for larger observation horizons (40 frames), the added complexity of the cross-stream residual connections yields the best performance, i.e., an accuracy of $90.3\%$ (see Table \ref{table:1}). Note that these results
clearly outperform previous results on the PREVENTION dataset \cite{Izquierdo2019}.

\begin{table}[h!]
\centering
\begin{tabular}{|c|c|c|c|c|c|} 
 \hline
 & & \multicolumn{4}{c|}{ROI size} \\
 \hline
 Method & Obs. Horizon & x1 & x2 & x3 & x4\\
 \hline
 Disjoint & 20 & 83.22 & 86.18 & 86.26 & 87.43 \\ 
 Disjoint & 30 & 83.55 & 86.69 & 86.84 & 86.68 \\
 Disjoint & 40 & 84.97 & 87.69 & \textbf{89.46} & 88.79 \\ 
 \hline
 ST & 20 & 83.39 & 85.03 & 86.51 & 86.16 \\ 
 ST & 30 & 84.38 & 84.70 & 85.36 & 84.73 \\
 ST & 40 &  86.02 & 87.83 & \textbf{90.30} & 89.64 \\ 
 \hline
 \end{tabular}
\caption{Disjoint Two-Stream Network and Spatiotemporal Multiplier Network Classification Accuracy ($\%$).}
\label{table:1}
\end{table}

\subsection{Lane change prediction results}
The ability of both methodologies to predict the future lane-change manoeuvre of surrounding vehicles is evaluated using an observation horizon of 20 frames (2 seconds) and prediction horizon (TTE) of 10 and 20 frames. The results for both approaches are depicted in Table \ref{table:2}.

As can be observed, surprisingly, the results for longer prediction horizons are better, i.e., the accuracy of both models for $TTE=20$ frames is approximately $5\%$ higher in all cases for both models than for a $TTE=10$. This can be partially explained by the complexity of the two-stream models that improves generalization with a more complex target to learn. The best accuracy for the disjoint two-stream network is given for a prediction horizon of 2 seconds and a ROI size of x3, yielding $91.02\%$. For the spatiotemporal multiplier network, the best prediction accuracy, $91.94\%$ is given for a TTE of 2 seconds and a ROI size of x4.

Up to our knowledge, these are the first prediction results so far using the PREVENTION dataset.

\begin{table}[h!]
\centering
\begin{tabular}{|c|c|c|c|c|c|}
 \hline
 & & \multicolumn{4}{c|}{ROI size} \\
 \hline
 Method & TTE & x1 & x2 & x3 & x4\\
 \hline
 Disjoint & 10 & 84.05 & 84.54 & 85.20 & 85.36 \\
 Disjoint & 20 & 85.20 & 88.82 & \textbf{91.02} & 90.92 \\ 
 \hline
 ST & 10 & 84.70 & 85.69 & 85.20 & 86.51 \\
 ST & 20 & 86.84 & 90.30 & 91.45 & \textbf{91.94} \\ 
\hline 
\end{tabular}
\caption{Disjoint Two-Stream Network and Spatiotemporal Multiplier Network Prediction Accuracy ($\%$). Observation horizon = 20.}
\label{table:2}
\end{table}


\section{Conclusions}
\label{sec:conclusions}
In this paper, two video action recognition approaches have been adapted, trained and evaluated to perform lane-change classification and prediction of surrounding vehicles in highway scenarios using the PREVENTION dataset. The problem was posed as an action recognition problem using visual cues from cameras, i.e., using the same source of information and approach that human drivers use to anticipate these maneuvers.

Both approaches, the disjoint two-stream convolutional network and the spatiotemporal multiplier network, are based on two different paths obtained from the same sequence of frames: a spatial stream in the form of individual region appearance, and a motion stream in the form of dense optical flow across frames. The complexity of the second model is based on the use of more complex architectures (ResNet50) and the use of multiplicative residual gating from the motion stream to the appearance stream.

Different ROI sizes have been evaluated, being the larger regions (x3 and x4) the ones providing the better results, due to the fact that they implicitly incorporate context information and iteration with other vehicles. The best configuration (spatiotemporal multiplier network with ROI size of x4 and observation horizon of 2 seconds) yields almost a $92\%$ of lane-change prediction accuracy two seconds earlier.

As future works, we plan to evaluate other action recognition approaches such as I3D models \cite{Carreira2017} and SlowFast Networks \cite{Feichtenhofer2019}.

\section*{ACKNOWLEDGMENT}

This work was supported in part by Spanish Ministry of Science, Innovation and Universities (Salvador de Madariaga Mobility Grant PRX18/00155 and Research Grant DPI2017-90035-R) in part by the Community Region of Madrid (Research Grant 2018/EMT-4362 SEGVAUTO 4.0-CM) and in part by the Air Force Office of Scientific Research USA (FA9550-18-1-0054) the Canada Research Chairs Program (950- 231659) and the Natural Sciences and Engineering Research Council of Canada (RGPIN-2016-05352).


\bibliographystyle{IEEEtran}
\bibliography{IEEEabrv,references}

\end{document}